\documentclass[journal]{IEEEtran}
\usepackage{amsmath,amsfonts}
\usepackage{array}
\usepackage[caption=false,font=normalsize,labelfont=sf,textfont=sf]{subfig}
\usepackage{textcomp}
\usepackage{stfloats}
\usepackage{url}
\usepackage{verbatim}
\usepackage{graphicx}
\usepackage{cite}

\usepackage{times}
\usepackage{epsfig}
\usepackage{graphicx}
\usepackage{amsmath}
\usepackage{amssymb}
\usepackage{booktabs}
\usepackage{xcolor}
\usepackage{comment}
\usepackage[table]{xcolor}
\usepackage{capt-of}
\usepackage[normalem]{ulem}
\usepackage{tabularx}
\usepackage[symbol]{footmisc}
\usepackage{multirow}
\usepackage{soul}
\usepackage[ruled]{algorithm2e}
\usepackage{caption}

\begin{document}

\title{NeRF-MIR: Towards High-Quality Restoration of Masked Images with Neural Radiance Fields}

\author{Xianliang Huang, 
Zhizhou Zhong,
Shuhang Chen,
Yi Xu, 
Jihong Guan$^*$,
Shuigeng Zhou$^*$, $Senior~Member$

\thanks{Xianliang Huang, 
Zhizhou Zhong,
Shuhang Chen,
Yi Xu, and
Shuigeng Zhou are with Shanghai Key Lab of Intelligent Information Processing, and the College of Computer Science and Artificial Intelligence, Fudan University, Shanghai 200433, China. \protect
E-mail: \{huangxl21, 22210240402, 22210240123, yxu17\}@m.fudan.edu.cn, sgzhou@fudan.edu.cn}  
\thanks{Jihong Guan is with the School of Computer Science and Technology, Tongji University, Shanghai 201804, China. \protect
E-mail: jhguan@tongji.edu.cn}
\thanks{$^*$Correspondence author.}

}



\maketitle

\begin{abstract}
Neural Radiance Fields (NeRF) have demonstrated remarkable performance in novel view synthesis. However, there is much improvement room on restoring 3D scenes based on NeRF from corrupted images, which are common in natural scene captures and can significantly impact the effectiveness of NeRF. This paper introduces NeRF-MIR, a novel neural rendering approach specifically proposed for the restoration of masked images, demonstrating the potential of NeRF in this domain. Recognizing that randomly emitting rays to pixels in NeRF may not effectively learn intricate image textures, we propose a \textbf{P}atch-based \textbf{E}ntropy for \textbf{R}ay \textbf{E}mitting (\textbf{PERE}) strategy to distribute emitted rays properly. This enables NeRF-MIR to fuse comprehensive information from images of different views. Additionally, we introduce a \textbf{P}rogressively \textbf{I}terative \textbf{RE}storation (\textbf{PIRE}) mechanism to restore the masked regions in a self-training process. Furthermore, we design a dynamically-weighted loss function that automatically recalibrates the loss weights for masked regions. As existing datasets do not support NeRF-based masked image restoration, we construct three masked datasets to simulate corrupted scenarios. Extensive experiments on real data and constructed datasets demonstrate the superiority of NeRF-MIR over its counterparts in masked image restoration. 
\end{abstract}


\section{Introduction}
\IEEEPARstart{N}{eural} Radiance Fields (NeRF) have recently demonstrated exceptional performance in novel view synthesis, which aims to reconstruct arbitrary views of a scene based on a sequence of images with known camera poses. Fundamentally, NeRF employs a multi-layer perceptron (MLP) to map 3D positions and 2D view directions to color and density. The output of MLP can subsequently be rendered into 2D images via volume rendering.
Up to now, NeRF has been applied to various scenarios, including city-scale scene representation~\cite{rematas2022urban,tancik2022block}, 3D-aware image synthesis~\cite{schwarz2020graf,niemeyer2021giraffe,xu2024mambatalk,mi2025data}, deformable 3D reconstruction~\cite{wang2022clip,guo2021ad,liu2021neural,park2021nerfies}
and pose estimation~\cite{wang2021nerf,jeong2021self,zhao2025analyzing} etc. Among them, several of these endeavors address traditional image processing tasks, including image editing~\cite{liu2021editing}, denoising~\cite{pearl2022nan} and deblurring~\cite{ma2022deblur}.
However, in the literature, there is a notable deficiency in the works on restoring NeRF-based masked images, which implies the potential of NeRF to effectively remove substantial distractors from images.

\begin{figure}
   \centering
   \includegraphics[width=\linewidth]{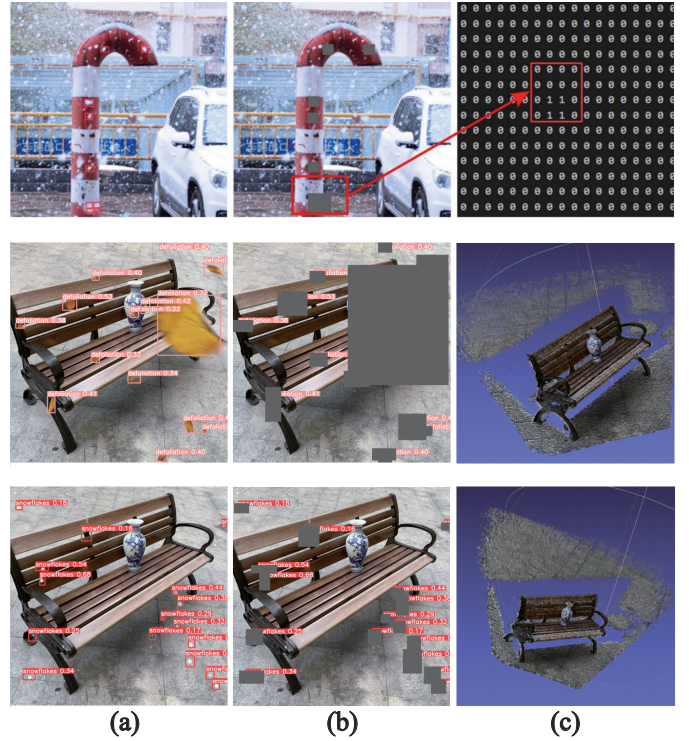}
   \caption{The process of obtaining the mask matrix is illustrated in the first row. Two examples of common image distractors in nature are shown in the second and third rows: (a) the output results of YOLOv5, (b) the masked results based on the center coordinates of bounding boxes, and (c) the reconstructed camera parameters of masked images.}
   \label{fig:process of mask matrix}
\end{figure}

A significant limitation of vanilla NeRF is its reliance on clean multi-view inputs. However, capturing images without multiview corruption is challenging due to the presence of various dynamic distractions in natural environments such as petals, snowflakes, and defoliation. These distractions contribute to quality degradation, reduced visual appeal, and the loss of valuable information. Essentially, the corrupted pixels are equal to masking out with a consistent RGB value, thus transforming the problem to masked image restoration. There are two reasons: (1) There is no one-size-fits-all solution for restoring various distractors in images. The presence of diverse distractors requires hand-crafted design features or well-designed models for the removal of each type of distractors. (2) Distractors are modeled by masked patches in captured multi-view images at random positions, reflecting the inherent random distribution characteristics of natural distractors in images. 
Actually, masked regions within images detract the primary objects, diminishing the overall image quality. Consequently, restoring masked images has the potential to enhance the accuracy and reliability of various computer vision tasks, including object recognition, 3D scene reconstruction, and image segmentation. 

To generalize the problem, the corrupted pixels can be masked with a fixed RGB value. Specifically, given multi-view corrupted images, we first annotate the positions of distractors in images manually or by target detection algorithms. Then, these regions are masked with a series of regular patches. The masked pixels in the patches are replaced with a default RGB value. Finally, these patches are transformed into a 0/1 mask matrix, where 1 means masked and 0 indicates non-masked. The process of imposing masked regions on images and establishing the corresponding mask matrices is illustrated and exemplified in Fig.~\ref{fig:process of mask matrix}. 
We model scene image interference using masked patches in captured images at random positions from different views because the positions of natural distractors in images are essentially random. Then, the problem turns to masked image restoration.

NeRF provides an effective approach for extracting consistent features from the underlying 3D scenes across different views. In light of this, we tackle the problem of restoring masked images by proposing a new method called NeRF-MIR. 
Specifically, recognizing the limitation of randomly emitting rays to pixels in NeRF cannot effectively learn complex textures in images. We develop the \textit{Patch Entropy based Ray Emitting} (PERE) strategy, which ensures the proper distribution of emitted rays, facilitating the fusion of comprehensive information from images captured in different views. Additionally, we design a \textit{Progressively Iterative Restoration} (PIRE) mechanism to restore masked images in a self-training manner. As the training process goes, the masked regions gradually reveal their content, and more attention should be paid on the losses associated with these regions. To this end, we introduce a dynamically weighted loss function that automatically adjusts the weights for masked and non-masked regions. The dynamically weighted loss has the advantage of plug-and-play with existing NeRF variants.

\begin{figure}[t]
  \centering
   \includegraphics[width=\linewidth]{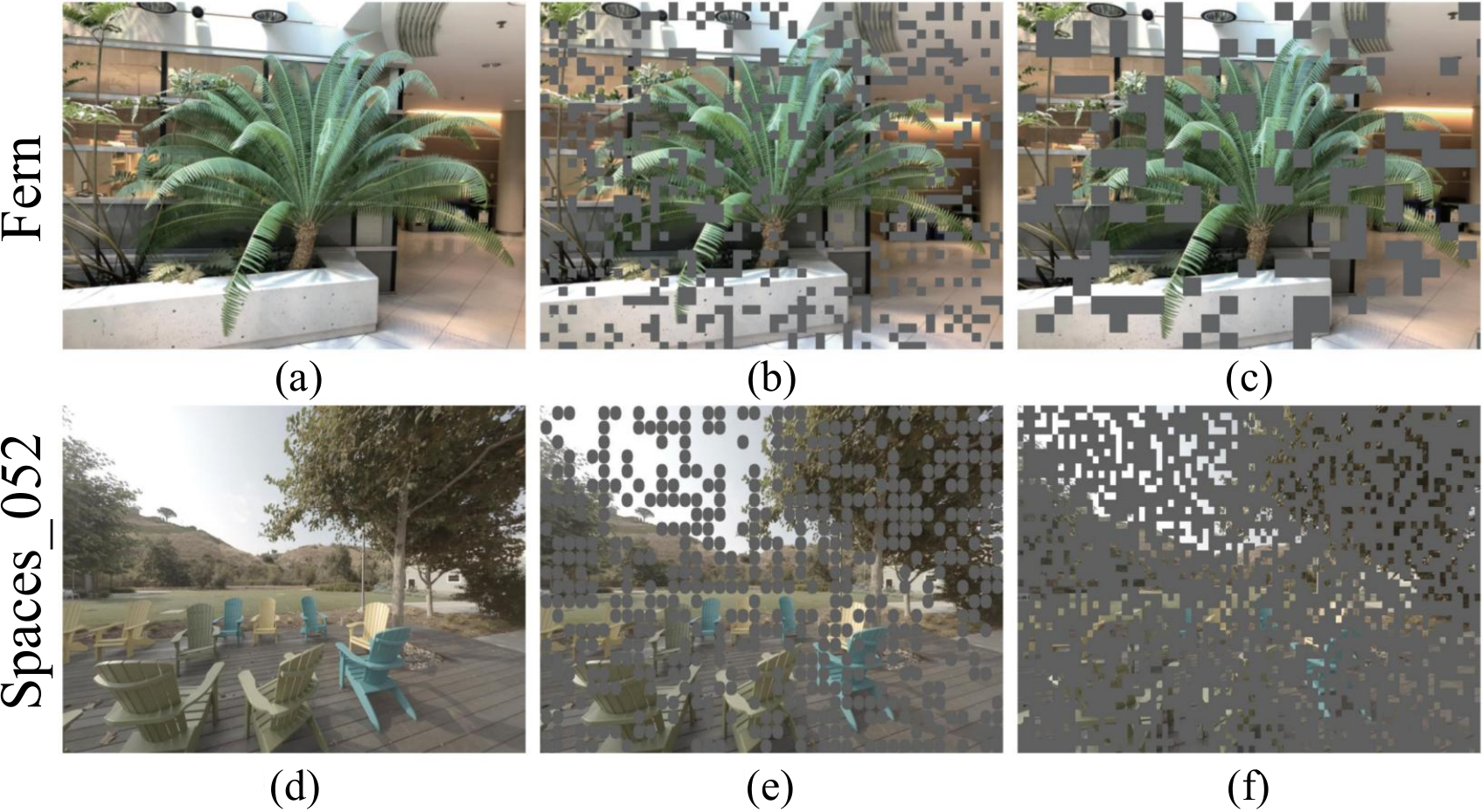}
   \caption{Sample scenes from our LLFF-M and Spaces-M datasets. The images of \emph{Fern} (a) and \emph{Spaces\_073} (d) are first split into small square patches. Then, we randomly or fixedly select patches in each view to mask by different shapes, sizes, and levels, which are shown in (b-c) and (e-f), respectively.}
   \label{fig:mask-operations}
\end{figure}

Furthermore, because existing datasets do not support NeRF-based masked image restoration, 
we construct three datasets for experiments, including the light-field-forward-facing masked images dataset (LLFF-M in short), the spaces masked image dataset (Spaces-M in short) and Blender-M. Sample scenes of these datasets are demonstrated in Fig.~\ref{fig:mask-operations}. These datasets are derived from the existing LLFF, Spaces, Blender datasets and simulated corrupted scenarios. 

In summary, the contributions of this paper are as follows: 
\begin{itemize}
\item We propose NeRF-MIR, a NeRF-based method tailored for masked image restoration. To the best of our knowledge, this is the first work that exploits NeRF to restore masked images, showcasing substantial potential in dealing with large amounts of distractors. 
\item We develop a patch-based entropy ray emitting~(PERE) strategy to effectively emit rays on masked regions, and design a progressively iterative restoration~(PIRE) mechanism to restore masked regions in an iterative self-training manner.
\item To support the training of NeRF-MIR, we propose a dynamically weighted loss to jointly optimize masked regions and non-masked regions in each stage. This enables NeRF-MIR to further distribute rays properly.
\item We conduct extensive experiments on three self-constructed datasets LLFF-M, Spaces-M and Blender-M, and some realistic datasets. Experimental results demonstrate that NeRF-MIR outperforms various self-defined NeRF methods in masked image restoration, and it is capable of restoring various distractors in realistic scenes.
\end{itemize}

The rest of the paper is organized as follows: we review related work in Sec.~\ref{sec:related-work}, and introduce the preliminaries of neural radiance fields and the problem definition in Sec.~\ref{sec:Preliminary}. Our method for masked image restoration is presented in Sec.~\ref{sec:method}. Experimental results, analysis, and discussions are given in Sec.~\ref{sec:exp}. 
Finally, we conclude the paper in Sec.~\ref{sec:conclusion}.

\begin{figure*}[t]
    \centering
    \includegraphics[width=\linewidth]{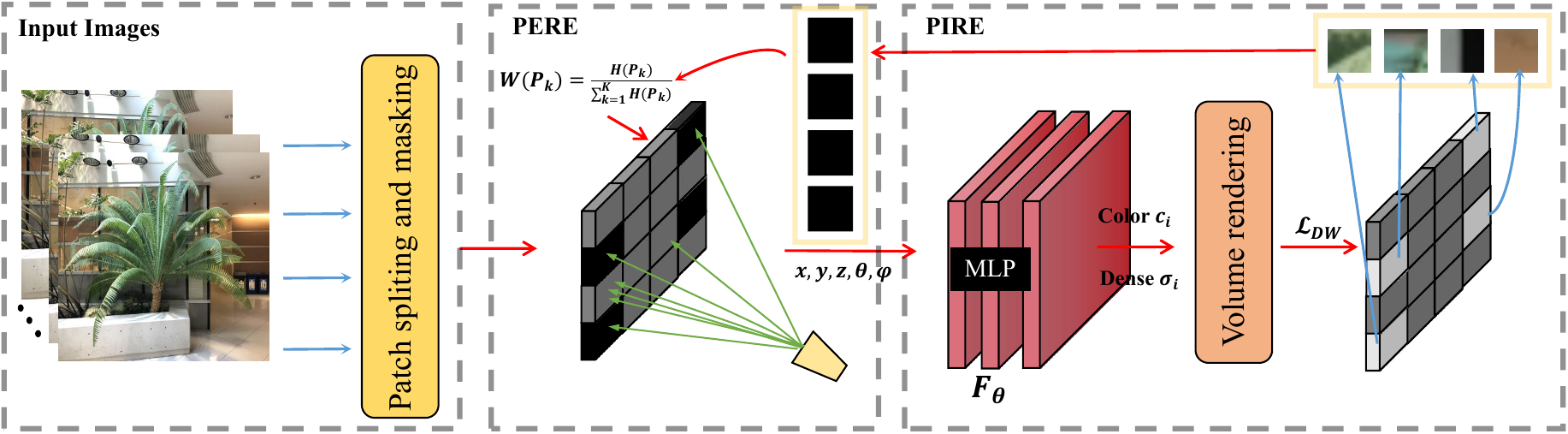}
    \caption{
    The framework of our method contains two major components: the Patch-Based Entropy for Ray Emitting (PERE in short) module and the Progressively Iterative Restoration~(PIRE in short) module. Input images are masked after patch segmentation, then we evaluate the patch entropy and redistribute rays to each patch as described in Sec.~\ref{sec:entropy}.
    We iteratively update the masked regions $\mathcal{P}_k$ by the PIRE mechanism that leverages multi-view information~(Sec.~\ref{sec:progressive}). The process is repeated several times, with the prediction of the prior stage serving as the training data for the next stage. }
    \label{fig:pipeline}
\end{figure*}

\section{Related Work}
\label{sec:related-work}
In this section, we review the related work from three aspects, including neural radiance fields (NeRF), NeRF-based image processing, image inpainting and scene restoration. 

\subsection{Neural Radiance Fields}
Neural radiance fields (NeRF) have made great progress in computer vision in the last two years. Unlike previous voxel, point cloud, and mesh-based methods, NeRF learns to represent the radiation fields in a 3D scene from a series of images by simply training a multi-layer perceptron~(MLP) network. The success of NeRF has inspired a variety of following works.
BRAF~\cite{lin2021barf}, NeRF$--$~\cite{wang2021nerf} and GNeRF~\cite{meng2021gnerf} try to train NeRF without known camera poses. NeRF-W~\cite{martin2021nerf} extends NeRF to model inconsistent appearance variations and transient objects across views successfully.
PixelNeRF~\cite{yu2021pixelnerf} and IBRNet~\cite{wang2021ibrnet} reconstruct a neural radiance field with a few images and achieve satisfactory performance.
Mip-NeRF~\cite{barron2021mip,barron2022mip} proposes to improve NeRF under multiscale and unbounded input, producing anti-aliased results.
Some works~\cite{schwarz2020graf,DBLP:conf/icmcs/ZhangYXGTW25} achieve great success in conditional generation by giving NeRF a random sampling of latent codes and poses.
Some other works~\cite{yu2021plenoctrees,garbin2021fastnerf,reiser2021kilonerf,muller2022instant} use baked or non-baked models to improve the training and inference speed of NeRF.

\subsection{NeRF-based Image Processing}
Neural radiance fields (NeRF) have made great progress in computer vision in the last two years. Here, we briefly review the latest advances in pose estimation~\cite{lin2021barf,wang2021nerf}, conditional generation~\cite{schwarz2020graf,cai2022pix2nerf}, better generalization~\cite{yu2021pixelnerf,wang2021ibrnet}, fast training and efficient inference~\cite{barron2021mip,yu2021plenoctrees,liang2024infnerf,DBLP:conf/cvpr/ZhangXRXTGWW25,muller2022instant}, which are related to our work. 

The advent of NeRF provides a new way to process images. Several works have tried to process images by implicit neural representation and achieve good results. For example, \cite{liu2021editing,kania2022conerf,wang2022clip} extend NeRF to edit the shape and appearance color, \cite{mildenhall2022nerf,huang2022hdr} enable users to perform high dynamic range (HDR) image view synthesis. Those works do not consider incomplete information from different perspectives~(e.g. noisy, blurry, and obstructions). NAN~\cite{pearl2022nan} is the first NeRF-based work that explicitly deals with significant photometric noise by leveraging inter-view and spatial information in images. Deblur-NeRF~\cite{ma2022deblur} investigates how to recover a sharp NeRF from blurry images. It proposes an approach to simulate the blurring process for reconstructing blurry views, making NeRF robust to blurry images. 


Some other works explore NeRF-based inpainting techniques, in this paper we use them as compared baselines to handle occlusions and missing regions in images. 
For instance, NeRF-In~\cite{shen2023nerf} leverages RGB-D priors for inpainting, enabling free-form region completion. Similarly, Spin-NeRF~\cite{mirzaei2023spin} integrates segmentation with inpainting to improve multi-view consistency, and NeRF-on-the-go~\cite{ren2024nerf} exploits uncertainty to filter distractors in the wild.
These methods mainly address structured degradations or rely on specific priors~(e.g. segmentable occlusions and uncertainty-guided filtering), whereas our approach directly tackles arbitrary random occlusions in real-world scenes.
Additionally, our work systematically explores the effect of training NeRF with masked images, a relatively unexplored direction. By focusing on random occlusions and proposing a dedicated NeRF-based restoration approach, we offer a novel solution distinct from existing inpainting and restoration frameworks.

\subsection{Image Inpainting and Scene Restoration}
Traditional image inpainting~\cite{bertalmio2000image} methods can be roughly divided into two categories: 1) Diffusion-based methods~\cite{ballester2001filling,bertalmio2003simultaneous,levin2003learning} fill the masked regions with neighboring content, which are prone to generate artifacts when the texture varies greatly; 2) Patch-based methods~\cite{barnes2009patchmatch,jia2004inference} fill the masked regions by searching patches of similar image features or from other reference images. They cannot generate new semantic content due to a lack of high-level feature understanding.
With the development of CNNs and GANs, learning-based methods~\cite{zhou2024seeing,liu2018image,yeh2017semantic} predict pixels in the masked regions in a semantics-aware manner. So they can produce high-quality textures and synthesize visually plausible content, especially for natural scenes~\cite{iizuka2017globally}.
The majority of prior works~\cite{shang2023deformable, zhang2023ps, cao2023zits++, zhang2024mmginpainting, zhou2025devil} on image inpainting mainly focus on individual 2D images to fill masked regions, they do not utilize 3D geometry information, and cannot generate sharp 3D dynamic scene videos from a series of masked images. While our method exploits NeRF and 3D consistency to infill masked, posed image collections.

In the restoration of disturbed scenes, typical distractors include snowflakes, leaves, etc. Traditional image processing methods~\cite{tian2018snowflake,barnum2010analysis} focus on designing hand-crafted features, which are tailored to the dynamic characteristics, sizes, and shapes of the distractors. Methods for rain removal usually cannot be directly applied to snowflake removal. For example, \cite{zhen2013new} and \cite{kim2015video} improve visibility but cannot remove snowflakes from a scene. Similarly, a 2D degraded image restoration method~\cite{zha2022low} based on image priors for denoising, has limited applicability to snowflake removal and inpainting tasks. 
Deep learning-based approaches~\cite{cai2016dehazenet,liu2018desnownet, zhou2025learning} are more effective than the previous hand-crafted features. They can handle only snowflakes or raindrops, but not both simultaneously. Furthermore, training such networks requires large amounts of labeled data, which are difficult to acquire.

Our NeRF-MIR is suitable for restoring masked images caused by different distractors. It is also able to recover 3D video scenes from several masked images of different views, which is different from the previous 2D to 2D works. In particular, given a set of posed RGB images paired with binary masks, our method learns to inpaint masked-out pixels or patches of pixels.

\section{Preliminary and Problem Definition}
\label{sec:Preliminary}
To represent 3D scenes, NeRF introduces a neural implicit function $F_{\Theta}: (\mathbf{p},\mathbf{v}) \rightarrow (\mathbf{c}, \sigma)$, where the 3D position $\mathbf{p} = (x,y,z)$ and the view direction $\mathbf{v} = (\theta, \phi)$ are taken as input to generate a volume density $\sigma$ and an RGB color $\mathbf{c} = (r, g, b)$. The mapping $F_{\Theta}$ is typically implemented by an MLP network, and $\Theta$ denotes the parameters of $F$. Theoretically, the rendered RGB color $C(\mathbf{r})$ can be calculated via integrating the predicted densities and colors along a ray $\mathbf{r}(t)=\mathbf{o}+t\mathbf{v}$ in NeRF. Due to the continuity of output value $(\mathbf{c}, \sigma)$ along the ray, the volume rendering integral equation~\cite{max1995optical} is numerically approximated by the following quadrature rule:
\begin{equation}\label{eq:ray_casting}
\hat{C}(\mathbf{r}) = \sum_{i=1}^{N}T_i(1-\exp(-\sigma_i\delta_i)) \mathbf{c}_i,
\end{equation}
where $\mathbf{r}$ denotes a ray, $N$ is the number of samples, and $\delta_i = t_{i+1} - t_{i}$ is the distance between the $i$-\text{th} point and its adjacent samples.
Note that $T_i$ is the accumulated transmittance, representing the probability along the ray without being intercepted until the $i$-\text{th} point. $T_i$ is formulated as follows: 
\begin{equation}
T_i = \sum_{j=1}^{i-1} \sigma_j \delta_j
\end{equation}

The points on the ray are sampled using a two-stage hierarchical technique for rendering efficiency. These points are sampled uniformly in the first stage, and the importance sampling is adopted in the second stage with the density predicted in the first stage.
Since the process is fully differentiable, a mean square loss function is used to update the MLP parameters $\Theta$ as follows:
\begin{equation}
\label{eq:nerf-loss}
    \mathcal{L}_{mse} =  \sum_{\mathbf{r}\in\mathcal{R}}\|(\hat{C}^c(\mathbf{r}) - C(\mathbf{r})) \|^2_2  \\ +  \|(\hat{C}^f(\mathbf{r}) - C_i(\mathbf{r}))\|^2_2,
\end{equation}
where $\mathcal{R}$ denotes a batch of rays. $C(\mathbf{r})$ is the ground truth, $\hat{C}^{c}(\mathbf{r})$ and $\hat{C}^f(\mathbf{r})$ are the coarse and fine volume predicted RGB colors of ray $\mathbf{r}$, respectively.
Besides, positional encoding is also employed before the MLP for mapping the input coordinates $(\textbf{p}, \textbf{v})$ to a higher dimensional space, which helps represent high-frequency scenes.


Given a set of 
 $N$ posed images $I$=$ \{I_i\}^{N}_{i=1} $, where the images do not strictly adhere to typical NeRF assumption of being captured from a perfectly 3D consistent and static scene. The images are contaminated by random occlusions \( M \), with the corresponding regions replaced by a fixed pixel value. The problem addressed in this paper is to reconstruct a complete 3D scene by training a clean radiance field \( F_\theta \) that restores the pixel values in the masked regions with the following objective function:
\begin{equation}
    \min_{\Theta} \mathcal{L}(F_\Theta, I, M),
\end{equation}
where \( \mathcal{L} \) is a loss function that measures the quality of restoration based on the mask regions $M$ and the contaminated images $I$.     

\section{Methodology}
\label{sec:method}
\subsection{Overview}
Our method fills in masked regions by repeatedly training a NeRF model and augmenting the input data with the model's intermediate predictions for masked region restoration.
The framework of NeRF-MIR depicted in Fig.~\ref{fig:pipeline} consists of two novel components: PERE and PIRE. In PERE, input images are partitioned into square patches, each of which is assigned a weight based on patch entropy. Patches with large weights are sampled more frequently, emphasizing intricate textures during training. PIRE facilitates iterative training, where predictions of masked pixels from the current stage inform subsequent training. PERE contributes to effective ray emission for intricate texture extraction and PIRE progressive enhancement of restoration quality by mitigating masked region effects. Additionally, we design a dynamically-weighted loss to further alleviate the incorrect guidance of masked regions in the early stages of training. 

In our task, the stochastic pixel-based ray emission approach in the vanilla NeRF inevitably leads to a significant number of rays falling within the masked regions, resulting in lots of emitted rays being wasted for invalid information extraction. Therefore, the development of an effective ray-emitting strategy is critical. We focus on emitting more rays to non-masked regions with intricate textures, avoiding the casting of rays directly to the masked regions. Furthermore, the laziness of neural networks makes it easy for NeRF to overfit and directly produce black dots in masked regions. To address the issue, we propose a progressively iterative restoration scheme that exploits the results from different perspectives in the current stage to update masked regions and achieve high-quality synthesis in the subsequent stage.

\begin{figure}
  \centering
   \includegraphics[width=\linewidth]{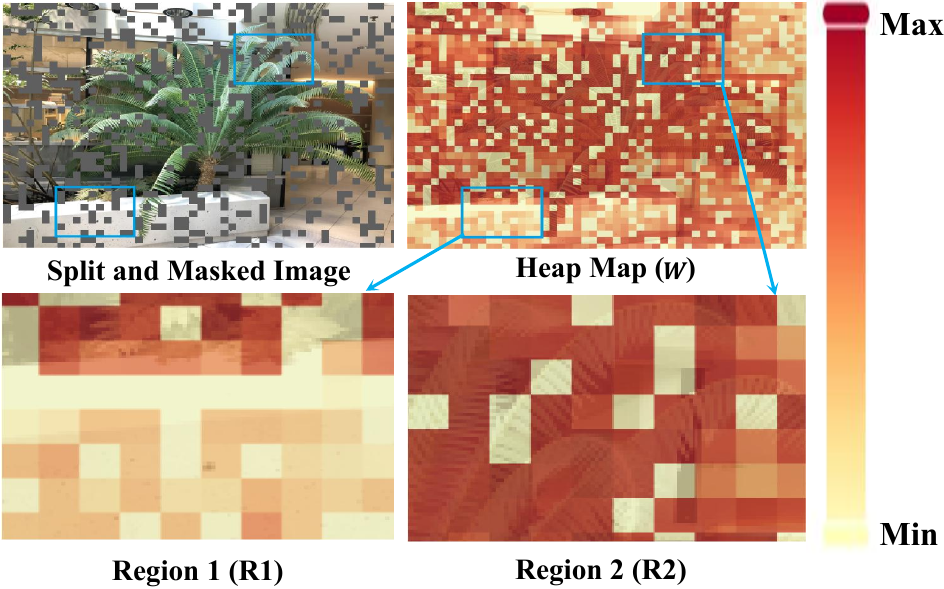}
   \caption{The heat map of ray distribution. The areas of smooth texture (R1) and intricate texture (R2) are bounded by a blue box.}
   \label{fig:entropy heatmap}
\end{figure}

\subsection{Patch-Based Entropy for Ray Emitting}\label{sec:entropy}
As it is difficult to evaluate the local detail information at a single pixel, the key idea is to emit rays based on patches and determine the number of rays according to the amount of detail information in each patch. We employ image entropy to quantitatively measure the richness of detailed information within each patch.

\noindent \textbf{Patch Splitting and Masking.}
Suppose we have a set of masked images $\mathcal{I} = \{I_1, I_2,...,I_{N}\}$ that the masked pixels are replaced with a default RGB value (96,96,96), each image $I_i \in \mathcal{I}$ is randomly scattered with a certain number of non-overlapping square masks. 
Taking the masked images as input, we divide them into gridded patches of size $l \times l$. Formally, we denote the indices of gridded patches in the image $I_i$ as $\mathcal{P}_i = \{P^i_k\ \mid 1 \le k \le K \} $. Here, $K=\frac{H\times W}{l\times l}$ is the number of patches in the image $I_i$, $H$ and $W$ are its height and width, respectively. By setting the masked level (the ratio of masked patches over all patches), mask size and shape, we randomly select patches and mask them.

\noindent \textbf{Calculation of patch entropy.}
Then, we denote the relative frequency of pixel value $m$ in the $k$-th patch as $p_k(m)$, which can be calculated as:
\begin{equation}
p_{k}(m)=\frac{1}{l\times l}\sum_{i=1}^{l\times l}\mathbb{I}(x_{n} = m),
\label{eq:color_density}
\end{equation}
where $x_n$ is the intensity of the $n$-th pixel in the patch.

Following the image entropy introduced in \cite{gull1984maximum}, we have the patch entropy of relative frequency as follows:
\begin{equation}
H({P}_{k}) = -\sum_{j=1}^3 \sum_{m=1}^{M} p^{(j)}_{k}(m)\log p^{(j)}_{k}(m) ,
\label{eq:patch_entropy}
\end{equation}
where $j$ is the number of image channels, $M$ is the maximum value of colors, $H(\cdot)$ is the patch entropy.

\noindent \textbf{Ray emission scheme.}
Intuitively, patches with detailed textures should be assigned more rays, while patches of smooth backgrounds should have fewer rays. Therefore, we propose to assign the number of rays for each patch based on the following information-density weight, which is evaluated as follows:
\begin{equation}
W({P}_{k}) = \frac{H({P}_{k})}{ \sum_{k=1}^{K} H({P}_{k}) } .
\label{eq:patch_weight}
\end{equation}

To guarantee that each patch can be cast at least one ray, we set a threshold $W_{min}$ for the weight $W(.)$. The number of rays assigned for patch $P_k$ is as follows:
\begin{equation}\label{eq:ray_assign}
\mathcal{N}_{k} =
\begin{cases}
1  & \text{if~} W({P}_{k}) <  W_{min} \\
W({P}_{k}) \times \mathcal{N}_{r}   & \text{otherwise}
\end{cases}
\end{equation}
where $\mathcal{N}_{r} = K \times \mathcal{N}_{p} $ denotes the assumed total number of rays, and $\mathcal{N}_{p}$ is the assumed averaged number of rays for each patch (used as a hyper-parameter).
Fig.~\ref{fig:entropy heatmap} shows an example of ray distribution in heat map form. Patches with larger weights are sampled more frequently during training. The weight of a patch is proportional to the entropy of its pixel intensities.

\begin{figure}
  \centering
  \includegraphics[width=\linewidth]{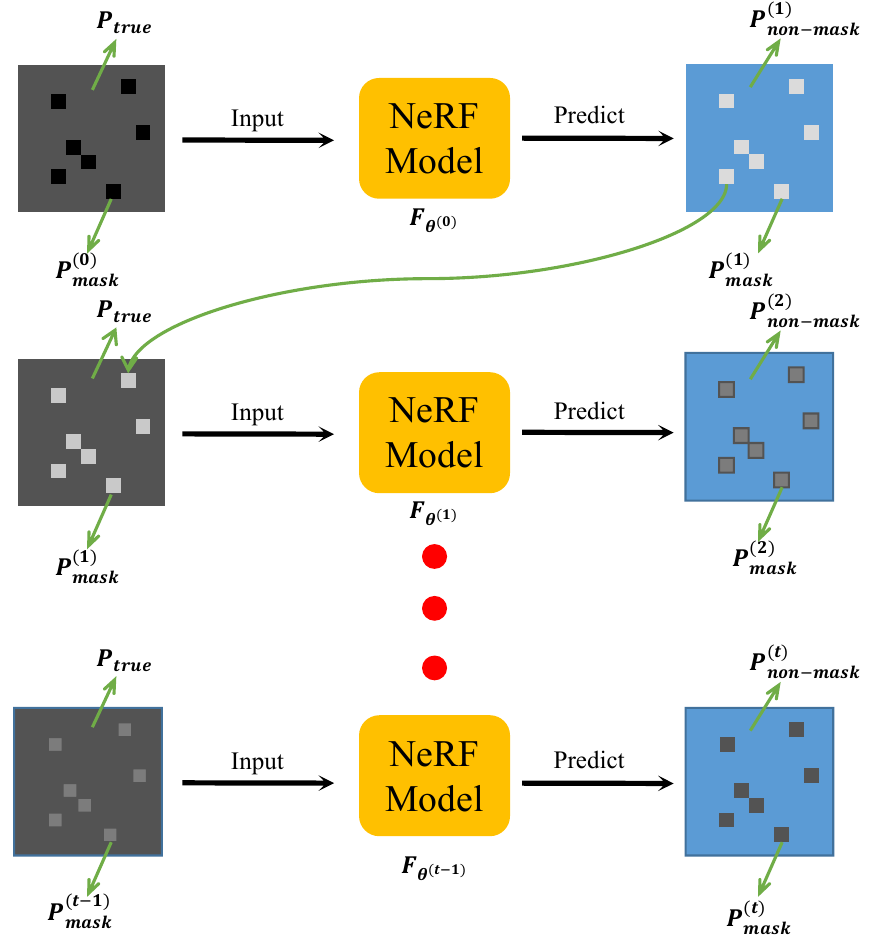}

  \caption{The detailed procedure of PIRE. The masked regions of input images in the $t$-th stage are replaced by the predictions from $\mathcal{P}^{(t-1)}_{mask}$ in the ($t$-1)-th stage. The predictions of each stage for masked-out pixels serve as additional training data for the following next stage.}
  \label{fig:progressive iteration}
\end{figure}

\begin{algorithm}[t]
	\caption{\textbf{Progressively Iterative Restoration (PIRE)}: 
    Consider mask patch set $\mathcal{P}_{mask}$ and non-mask patch set $\mathcal{P}_{non-mask}$ split from corrupted image $I_i$.
 }
        \textbf{Initialization:} Camera intrinsic $\mathbf{K}$, camera pose $\xi$ and patch splitting and masking pattern $\mathbf{v}$.
        
        \KwIn{Corrupted images ${I_i}$, total training epochs ${T}$, iterative stage number ${t}$, dynamic weight parameter $\alpha^{(1)}$.}
        
	\KwOut{Output clean images ${\hat{I_i}}$ and NeRF model ${F}_{\theta^{(t)}}$ with parameter $\theta^{(t)}$.}
        $\mathcal{P}_{true},\mathcal{P}^{(0)}_{mask} \leftarrow \mathbf{v}(I_i)$
        
	\For {$j$ from 1 to t}{   
            \textbf{repeat} $[T/t]$ times:

            \ \ \ \ \  take gradient descent step
            
            \ \ \ \ \  $\mathcal{L}_{DW} \leftarrow \nabla_{\theta}\theta^{(j-1)}(\mathcal{P}_{true},\mathcal{P}^{(j-1)}_{mask}, \alpha^{(j-1)})$
            
        $[\mathcal{P}^{(j)}_{non-mask},\mathcal{P}^{(j)}_{mask}] \leftarrow {F}_{\theta^{(j-1)}}( \mathcal{P}^{(j-1)}_{mask}, \mathbf{K}, \xi)$
            

        


        $\alpha^{(j)} = \alpha^{(j-1)} + \Delta\delta$
        
        $\hat{I_i} \leftarrow [\mathcal{P}_{true},\mathcal{P}^{(j)}_{mask}]$
        
	}
	\Return ${\hat{I}_i}$, ${F}_{\theta^{(t)}}$
\label{algo5_v1}
\end{algorithm}

\subsection{Progressively Iterative Restoration}\label{sec:progressive}
In our preliminary experiments, we observed that the results of using only PERE were not satisfactory. From the results of NeRF+PERE, we can see that quite a lot of masked areas (small grey blocks) are not restored well: either partially restored or remaining unrestored.

Therefore, we propose a self-training-like approach termed \textit{progressively iterative restoration} (PIRE) to solve this problem. The detailed procedure is shown in Fig.~\ref{fig:progressive iteration}.
Let $\mathcal{P}^{(0)}_{mask}$ and $\mathcal{P}_{true}$ represent the initial masked and non-mask regions, respectively. Suppose the total number of training epochs is $T$, we divide $T$ equally into $t$ stages for progressive refinement.
In the initial training epoch of the $t$-th stage, we can formulate the prediction with model parameter $\theta^{(t-1)}$ as follows:
\begin{equation}\label{eq:progressive}
[\mathcal{P}^{(t)}_{non-mask},\mathcal{P}^{(t)}_{mask}]={F}_{\theta^{(t-1)}}(\mathcal{P}_{true},\mathcal{P}^{(t-1)}_{mask})
\end{equation}
{Then, we replace the masked regions with their corresponding predictions $\mathcal{P}^{(t)}_{mask}$ and derive a new model ${F}_{\theta^{(t)}}$ with $[\mathcal{P}_{true},\mathcal{P}^{(t)}_{mask}]$ in the next stage.} Algorithm \ref{algo5_v1} outlines the process of PIRE in our NeRF-MIR.
In short, the core idea of PIRE is to leverage the rendering results of the NeRF model ${F}_{\theta^{(t-1)}}$ in the $(t-1)$-th stage as initial values of masked patches for training the model ${F}_{\theta^{(t)}}$ in the $t$-th stage.
By PIRE, each of the masked regions converges to a high-quality restoration status, which is close to the result of the vanilla NeRF on the non-masked data. Though PIRE shares some similarity with transductive learning in terms of leveraging unlabeled data for self-improvement in specific cases, our PIRE focuses on utilizing multi-view information to predict unlabeled patches and perform self-training.

\begin{figure*}[t]
    \centering
    \includegraphics[width=\linewidth]{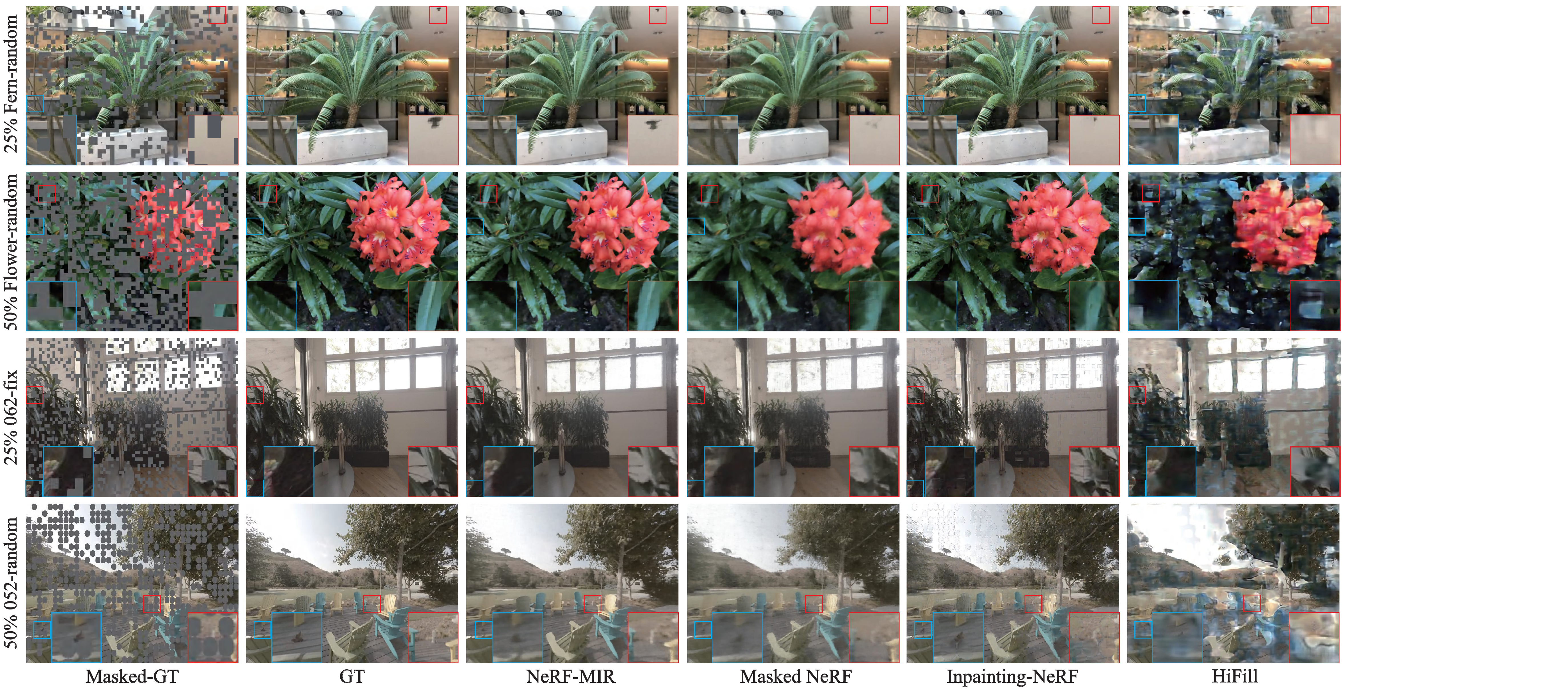}
    \caption{Qualitative comparison on \emph{Fern}, \emph{Flower}, \emph{Spaces\_062} and \emph{Spaces\_052} scenes from the LLFF-M and Spaces-M datasets. Given the input images (GTs) and corresponding masked images (Masked-GTs), we visualize the output results of HiFill, Inpainting-NeRF, Masked NeRF and ours in a consistent view. There are four different masking settings from top to bottom: 25\%-random square, 50\%-random square, 25\%-fixed square, and 50\%-random round. The compared regions are enlarged in red and blue boxes.}
    \label{fig:Results on LLFF-M}
\end{figure*}


\subsection{Dynamically-weighted Loss}
{In the initial stage, since the content of masked regions is unknown, losses associated with these areas will not be considered. As the training goes on, the masked areas are gradually restored, and more attention should be paid to the losses associated with these regions. Consequently, we introduce a dynamic weight parameter, denoted as $\alpha$, which evolves with the training process. The dynamically weighted loss for training NeRF-MIR, which adaptively adjusts the loss weight between masked and non-masked regions, is formulated as follows:}
\begin{equation}\label{eq:seen_ray_loss}
    \begin{gathered}
    \mathcal{L}_{DW} = (1-\alpha)(\sum_{\mathbf{r}\in\mathcal{R}}\|(\hat{C}(\mathbf{r}) - C(\mathbf{r})) \|^2_2)  \\ +  \alpha(\sum_{\mathbf{r}\in\mathcal{M}}\|(\hat{C}(\mathbf{r}) - C_i(\mathbf{r}))\|^2_2),
    \end{gathered}
\end{equation}
where $\mathcal{R}$ denotes a batch of rays emitting to unmasked regions and $\mathcal{M}$ denotes a batch of rays emitting to masked regions. {Rather than training NeRF from scratch in each stage, the preceding stage's model is important for refining the next stage's model. 
We initialize $\alpha$ to zero and increase it linearly because the mask regions gradually obtain the correct information from other views as the training process goes.}

The dynamically-weighted loss function follows an intuitive principle inspired by curriculum learning~\cite{bengio2009curriculum}, where the model gradually evolves from coarse to fine learning. This is implemented in our approach by progressively increasing the weight of the masked region loss as the iteration goes, allowing the model first to learn robust representations from unmasked regions, and then to refine its understanding of the occluded areas. This strategy aligns with the self-supervised learning principles~\cite{jaiswal2020survey}, where gradual weight adjustment leads to stable convergence and improved generalization.

The completion of the masked regions by PIRE allows PERE to distribute rays more rationally. While the distribution of rays by PERE facilitates the PIRE to restore the images better (with fewer artifacts in masked regions).

\begin{figure*}[t]
  \centering
   \includegraphics[width=\linewidth]{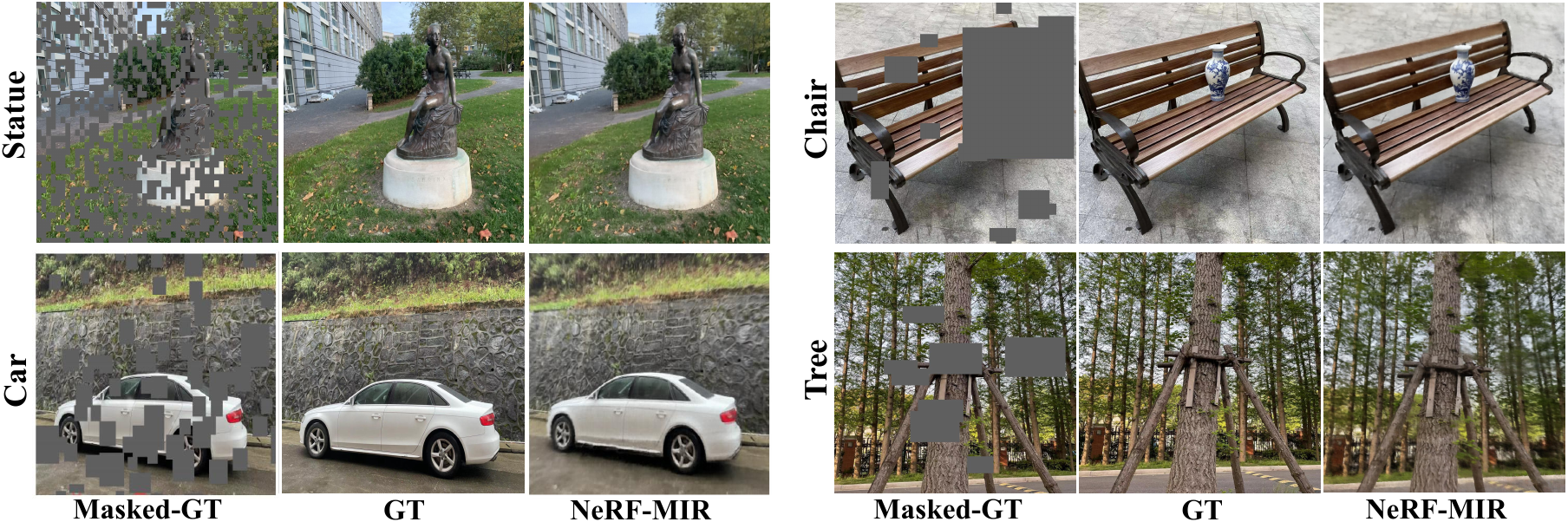}
   \caption{Qualitative results of NeRF-MIR on four different real-world scenes with snowflakes and defoliation as distractors. Multi-views are captured by iPhone 12 and YOLOv5 is adopted to detect snowflakes and defoliation. We mask the detected bounding boxes with default RGB values and visualize selected views in synthetic results. }
   \label{fig:realistic_result}
\end{figure*}

\begin{table*}[t]
\caption{Quantitative results on the LLFF-M and Spaces-M datasets. We test each setting five times with the same parameters and evaluate the average score. The best results are presented in bold. }

\label{tab:Quantitative results}
\centering
\footnotesize
\setlength{\tabcolsep}{1pt}
\newcolumntype{Y}{>{\centering\arraybackslash}X}
\begin{tabularx}{0.999\linewidth}{l||YYY|YYY|YYY|YYY}
\toprule
& \multicolumn{3}{c}{\scshape 25\% Masked Fern}
 & \multicolumn{3}{c}{\scshape 50\% Masked Flower}
 & \multicolumn{3}{c}{\scshape 25\% Masked Spaces\_062}
 & \multicolumn{3}{c}{\scshape 50\% Masked Spaces\_052}
\\
  \textsl{Method} & \multicolumn{1}{c}{\scriptsize PSNR$\uparrow$}    & \multicolumn{1}{c}{\scriptsize SSIM$\uparrow$} & \multicolumn{1}{c}{\scriptsize LPIPS$\downarrow$} & \multicolumn{1}{c}{\scriptsize PSNR$\uparrow$} & \multicolumn{1}{c}{\scriptsize SSIM$\uparrow$} & \multicolumn{1}{c}{\scriptsize LPIPS$\downarrow$} & \multicolumn{1}{c}{\scriptsize PSNR$\uparrow$} & \multicolumn{1}{c}{\scriptsize SSIM$\uparrow$} & \multicolumn{1}{c}{\scriptsize LPIPS$\downarrow$} & \multicolumn{1}{c}{\scriptsize PSNR$\uparrow$}     & \multicolumn{1}{c}{\scriptsize SSIM$\uparrow$} & \multicolumn{1}{c}{\scriptsize LPIPS$\downarrow$}   \\
  \midrule
\scriptsize \textit{HiFill}               & 20.32       & 0.77    & 0.35     &16.22      &0.53      &0.52   &21.87        &0.83    &0.45  &18.78     &0.70     &0.43              \\
\scriptsize \textit{Zits++}    & 21.88     & 0.82    & 0.26    & 18.77     & 0.68    & 0.41     & 15.33     & 0.71    & 0.42    & 15.33     & 0.69    & 0.42          \\
 \scriptsize \textit{Inpainting-NeRF}               & 24.95       & 0.85    & 0.26     & 23.32    & 0.79    & 0.31     & 28.28   & 0.91    & 0.33     & 24.26        & 0.79  & 0.41          \\
 
 \scriptsize \textit{Masked NeRF}               & 25.72       & 0.86   & 0.21     & 26.78    & 0.90    & 0.16     & 29.08        & 0.92    & 0.26   & 25.10   & 0.82    & 0.35            \\

\scriptsize \textit{NeRF-In}               & 25.13 & 0.86 & 0.23 & 24.32 & 0.84 & 0.28 & 27.94 & 0.90 & 0.30 & 24.87 & 0.80 & 0.38          \\

\scriptsize \textit{SPIn-NeRF}               & 26.04 & 0.88 & 0.21 & 25.16 & 0.87 & 0.25 & 28.67 & 0.91 & 0.28 & 25.43 & 0.81 & 0.36          \\

\scriptsize \textit{NeRF On-the-go}            & 27.32 & 0.90 & 0.19 & 26.51 & 0.89 & 0.22 & 30.10 & 0.93 & 0.26 & 26.02 & 0.84 & 0.33         \\
\scriptsize \textit{NeRF-MIR(ours)}              & \textbf{28.13}       & \textbf{0.92}     & \textbf{0.14}     & \textbf{29.18}    & \textbf{0.92}    & \textbf{0.13}    & \textbf{32.13}        & \textbf{0.95}    & \textbf{0.23} & \textbf{27.65}   & \textbf{0.86}    & \textbf{0.31}              \\
\midrule
\bottomrule
\end{tabularx}
\end{table*}




\section{Experiments}
\label{sec:exp}
In this section, we perform extensive experiments with synthetic and real scenes to validate the robustness and effectiveness of NeRF-MIR under the settings where the distractors are masked. Several baselines are compared with our method and visual results are presented. We also conduct ablation studies on key components and various settings. 

\subsection{Experimental settings}
\noindent \textbf{Baselines.}
We compare our method NeRF-MIR with both learning-based inpainting methods, including ZITS++~\cite{cao2023zits++}, LaMa~\cite{suvorov2022resolution}, TFill~\cite{zheng2022bridging} and HiFill~\cite{zeng2021cr}, and some recent NeRF-based approaches, including SPIn-NeRF~\cite{mirzaei2023spin}, NeRF-In~\cite{shen2023nerf}, and NeRF on-the-go~\cite{ren2024nerf}.
Furthermore, considering that this work is to restore 3D scenes from random-masked images by NeRF, we also propose the following baseline methods for fair comparisons: (1) A NeRF-based method that avoids the masked regions, which is denoted as \textbf{Masked NeRF}. (2) A NeRF-based method that infills the masked regions with 2D inpainting and is trained on vanilla NeRF, which is denoted as \textbf{Inpainting-NeRF}. Note that latent diffusion model~\cite{rombach2022high} is adopted as the inpainting backbone in Inpainting-NeRF. 


\noindent \textbf{Metrics.}
we employ peak signal-to-noise ratio (PSNR), structural similarity (SSIM)~\cite{wang2004image} and learned perceptual image patch similarity (LPIPS)~\cite{zhang2018unreasonable} as standard image quality metrics for the masked image restoration task to evaluate our NeRF-MIR, i.e., evaluating the quality of masked regions and the results of novel view synthesis.

\noindent \textbf{Training.}
We implement our NeRF-MIR in PyTorch \cite{misc_nerfpytorch} and
Python 3.9. The mask unit and patch size are set to $10 \times 10$ pixels. In each epoch, 1938 patches are randomly selected from the training images to emit an equal number of rays, with $\mathcal{N}_{p}$=2. 
We train each scene for 50k iterations, which takes around 10 hours on a single GeForce RTX 3090 GPU. The Adam optimizer is used~\cite{misc_adam} with default parameters. We schedule the learning rate to $5\times 10^{-4}$ at starting and decay it exponentially to $8\times 10^{-5}$. $t$ in PIRE is set to 5 and $\alpha$ is increased by 0.125 every 10000 epochs. We use an MLP with $4$ fully connected hidden layers, each layer having $64$ channels and ReLU activations.
For a fair comparison, $\mathcal{N}_{p}$ is set to ensure that the training images of LLFF-M and Spaces-M dataset emit an equal number of rays in proposed baselines.
On the other hand, we capture {20 $\sim$ 30 images} in realistic scenes by iPhone 12 with a resolution of $1080 \times 1440$. The \emph{car} and \emph{chair} scenes present in the main paper are trained for 100k and 80k iterations, respectively. 2000 patches are selected in each epoch to emit an equal number of rays. 

\noindent \textbf{Datasets.}
To evaluate the restoring result of masked images, we adopt {20 $\sim$ 40 images} from the forward-facing dataset LLFF~\cite{mildenhall2019local}, Spaces dataset~\cite{flynn2019deepview}, Blender dataset and the realistic dataset.
LLFF consists of 8 scenes and each scene contains 20-62 images with a resolution of $504 \times 378$. 
Spaces dataset~\cite{flynn2019deepview} uses 16 forward-facing cameras on a fixed rig, which contains 7-16 images for each scene with a resolution of $800 \times 480$. The realistic dataset contains 10 real-world scenes with $2268 \times 4032$ pixels captured in SPIn-NeRF~\cite{mirzaei2023spin} and our realistic scenes captured by Apple iPhone 12 and each image is $1276 \times 1276$ pixels in size.
These datasets lack images with masked areas. Thus, we build new datasets by splitting and masking intricate scenes from the aforementioned datasets. We denote these datasets as LLFF-M, Spaces-M and Blender-M, which consider different masked levels (the proportion of masked areas over the whole image), mask sizes and shapes (e.g. square and round). {In addition, we consider two masking styles: (1) \textit{random}: randomly masking the areas of each view image; (2) \textit{fixed}: masking the fixed areas of each view image. We use ``50\%-random square'' to indicate that the images are masked randomly in square shape over 50\% areas. Note that square masks can be combined with rectangular blocks of any size to fit distractors of different sizes.}  
Fig.~\ref{fig:mask-operations} illustrates some samples from our LLFF-M and Space-M datasets in different settings.

\subsection{Experimental Results}
\noindent \textbf{Results on LLFF-M and Spaces-M.} In this part, we first qualitatively compare our method with our self-constructed baselines ---Masked NeRF, Inpainting-NeRF, and HiFill~\cite{zeng2021cr}. Then, we quantitatively compare our method with four recent state-of-the-art approaches, including SPIn-NeRF, ZITS++, NeRF-In, and NeRF on-the-go. In the experiments, 
the masked regions within the input views are selected randomly for the \emph{Fern} scene and fixedly for the \emph{Spaces\_062} scene. We increase the masked levels to 50\% for \emph{Flower} and \emph{Spaces\_052} scenes and change the shape of masks.

Fig.~\ref{fig:Results on LLFF-M} show the 
qualitative results. We can see that our NeRF-MIR outperforms HiFill in restoring masked images. In comparison with NeRF-based approaches, NeRF-MIR closely resembles the ground truth in detailed synthesis. This success can be attributed to the effectiveness of PERE, which significantly reduces the number of rays reaching the masked regions. Simultaneously, PIRE alleviates the effect of masked regions, fully harnessing NeRF's capacity to synthesize visual-plausible results. Although the performance of Masked NeRF is close to that of NeRF-MIR, its strategy of attempting to avoid the masked regions completely leads to the omission of detailed information, e.g. the disappearing trunk and incomplete alarm in the red and blue boxes of \emph{Fern scene} in Fig.~\ref{fig:Results on LLFF-M}. In contrast, our PERE is based on the distribution of image entropy to automatically adjust the ray distribution, providing more flexible and robust results.

Tab.~\ref{tab:Quantitative results} presents the
Quantitative results, which demonstrate the superior performance of NeRF-MIR in image restoration. 
Comparing with learning-based method, our approach achieves significant improvement in PSNR, concretely, an average of 12.7\% and 9.4\% gain over Inpainting-NeRF and masked NeRF in 25\% LLFF-M. Similar improvements are also observed in 25\% Space-M, 50\% LLFF-M and 50\% Space-M. HiFill and ZITS++ perform worst in terms of PSNR, SSIM and LPIPS on all scenes, likely due to their inpainting-based nature, which struggles to maintain structural consistency in 3D scenes. 
While comparing with the recent NeRF-based methods, our NeRF-MIR obviously performs better than NeRF-In, SPIn-NeRF and NeRF on-the-go, demonstrating its excellent reconstruction quality in terms of PSNR, SSIM, and LPIPS, which highlights its strong ability to handle masked regions. 

\begin{figure}
   \centering
   \includegraphics[width=\linewidth]{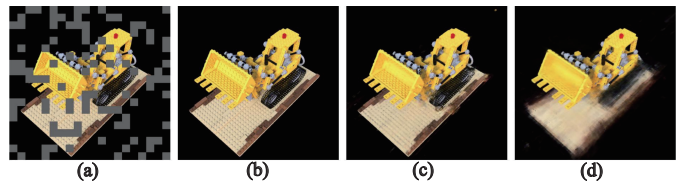}
   \caption{Qualitative results on \emph{lego} with 25\%-random square. We mask the input images (b) with patch size $40 \times 40$ and obtain the corresponding masked images (a). Then we select 800 patches for each epoch because the resolution of masked images is $800 \times 800$; (c) is the rendering result with few consistent views(9 images) as training images ; (d) is the rendering result with full views as training images.}
   \label{fig:lego_result}
\end{figure}

\noindent \textbf{Experiments on real-world scenarios and Blender-M.}
Here we further evaluate our method on real-world distractors in four different scenes. The snowflakes in \emph{Statue} and \emph{Car} scenes are replaced with default masks, and so are the defoliation as distractors in \emph{Chair} and \emph{Tree} scenes. Note that we utilize YOLOv5 to detect these distractors and the masked regions are calculated according to the output bounding boxes. The qualitative restoration results of NeRF-MIR are demonstrated in Fig.~\ref{fig:realistic_result}. For scenes with defoliation as distractors, our method successfully removes the huge defoliation and gets plausible visual restorations. In the other scene, snowflakes are barely visible in Fig.~\ref{fig:realistic_result}. Our method achieves excellent results in removing large amounts of snowflakes in real-world scenes.

On the other hand, we first train the NeRF model with all \emph{lego} images, but the experiments demonstrate ambiguous results in Fig.~\ref{fig:lego_result} (d). We assume that this is because the number of training epochs on the masked dataset is too small to support NeRF modeling 360° perspectives. Furthermore, there are large invalid regions on \emph{lego} dataset. Patches extracted from these areas do not contribute to the NeRF model restoring masked regions, which is the difference between the Blender dataset and the LLFF dataset. We attain advanced results in Fig.~\ref{fig:lego_result} (c) with only 9 views as training images that are relatively close to each other from \emph{lego} with 50000 epochs.

\begin{figure}
   \centering
   \includegraphics[width=\linewidth]{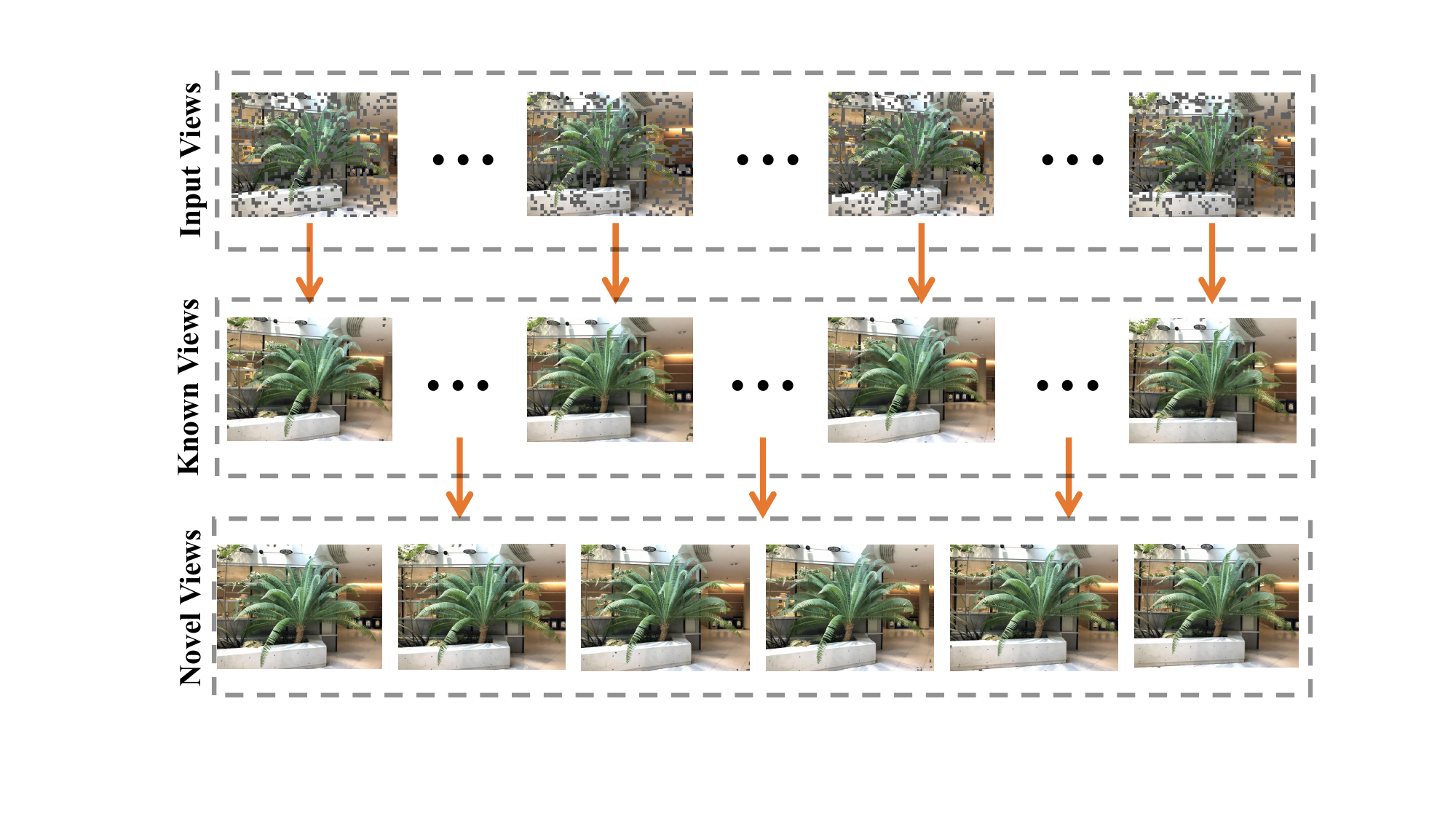}
   \caption{Multi-view results of the \emph{Fern} scene from the LLFF-M dataset with 25\% masked level. The novel views are extracted from the synthetic video at similar intervals. For an \textit{input view} with masks, the corresponding view w/o masks is called a \textit{known view}.}
   \label{fig:Multi-view results}
\end{figure}

\begin{table*}[t]
\caption{Quantitative results compared with NeRF on \emph{Fern}, \emph{Flower}, \emph{Fortress} and \emph{Room} scenes of LLFF dataset. Our NeRF-MIR achieves comparable results on PSNR, SSIM, and LPIPS.}

\label{tab:Comparison with NeRF}
\centering
\footnotesize
\setlength{\tabcolsep}{1pt}
\newcolumntype{Y}{>{\centering\arraybackslash}X}
\begin{tabularx}{0.999\linewidth}{l||YYY|YYY|YYY|YYY}
\toprule
& \multicolumn{3}{c}{\scshape \emph{Fern} scene of LLFF}
 & \multicolumn{3}{c}{\scshape \emph{Flower} scene of LLFF}
 & \multicolumn{3}{c}{\scshape \emph{Fortress} scene of LLFF}
 & \multicolumn{3}{c}{\scshape \emph{Room} scene of LLFF}
\\
  & \multicolumn{1}{c}{\scriptsize PSNR$\uparrow$}    & \multicolumn{1}{c}{\scriptsize SSIM$\uparrow$} & \multicolumn{1}{c}{\scriptsize LPIPS$\downarrow$} & \multicolumn{1}{c}{\scriptsize PSNR$\uparrow$} & \multicolumn{1}{c}{\scriptsize SSIM$\uparrow$} & \multicolumn{1}{c}{\scriptsize LPIPS$\downarrow$} & \multicolumn{1}{c}{\scriptsize PSNR$\uparrow$}    & \multicolumn{1}{c}{\scriptsize SSIM$\uparrow$} & \multicolumn{1}{c}{\scriptsize LPIPS$\downarrow$} & \multicolumn{1}{c}{\scriptsize PSNR$\uparrow$} & \multicolumn{1}{c}{\scriptsize SSIM$\uparrow$} & \multicolumn{1}{c}{\scriptsize LPIPS$\downarrow$}   \\
  \midrule
\scriptsize \textit{NeRF}               &28.97    & 0.93  & 0.11     &  30.27   & 0.95    & 0.08    &33.34     & 0.96   & 0.07     & 36.82  & 0.99    & 0.06    \\
\scriptsize \textit{NeRF-MIR}        &29.06    & 0.93  & 0.12     &  30.24   & 0.95    & 0.08     & 33.25     & 0.96   & 0.07     &37.0  & 0.99    & 0.06    \\

\toprule
& \multicolumn{3}{c}{\scshape \emph{Fern} scene of LLFF-M }
 & \multicolumn{3}{c}{\scshape \emph{Flower} scene of LLFF-M}
 & \multicolumn{3}{c}{\scshape \emph{Fortress} scene of LLFF-M}
 & \multicolumn{3}{c}{\scshape \emph{Room} scene of LLFF-M}
\\
& \multicolumn{1}{c}{\scriptsize PSNR$\uparrow$}    & \multicolumn{1}{c}{\scriptsize SSIM$\uparrow$} & \multicolumn{1}{c}{\scriptsize LPIPS$\downarrow$} & \multicolumn{1}{c}{\scriptsize PSNR$\uparrow$} & \multicolumn{1}{c}{\scriptsize SSIM$\uparrow$} & \multicolumn{1}{c}{\scriptsize LPIPS$\downarrow$}  & \multicolumn{1}{c}{\scriptsize PSNR$\uparrow$}    & \multicolumn{1}{c}{\scriptsize SSIM$\uparrow$} & \multicolumn{1}{c}{\scriptsize LPIPS$\downarrow$} & \multicolumn{1}{c}{\scriptsize PSNR$\uparrow$} & \multicolumn{1}{c}{\scriptsize SSIM$\uparrow$} & \multicolumn{1}{c}{\scriptsize LPIPS$\downarrow$}  \\
  \midrule

\scriptsize \textit{NeRF-MIR}              &28.13    & 0.92  & 0.14     &  29.68   & 0.93    & 0.10     & 32.92    & 0.96   & 0.08    & 33.86  & 0.98    & 0.07   \\

\midrule
\bottomrule
\end{tabularx}
\end{table*}


\noindent \textbf{Compared with NeRF on uncorrupted scenes.}
We also conduct NeRF-MIR on uncorrupted scenes, and the synthesized results of vanilla NeRF and NeRF-MIR on four scenes of LLFF dataset are presented in Tab.~\ref{tab:Comparison with NeRF}. Note that PIRE does not operate in scenes without masks. NeRF-MIR achieves comparable results to vanilla NeRF, indicating that our method does not deteriorate images without masks. 

\noindent \textbf{Novel view results.}
Our method is not only able to restore masked images but also able to generate {high-quality videos of dynamic scenes} from input masked images. 
We demonstrate the generated novel views of the \emph{Fern} scene, the results are shown in Fig.~\ref{fig:Multi-view results}. Though the other NeRF-based methods are also able to produce novel-view images, they cannot mitigate the effect of masked regions well, consequently producing relatively poor-quality results for all views. 
When the input views are contaminated with {masked regions}, our method outputs exceptionally photo-realistic results for unseen views, which are impossible for the existing image inpainting methods to synthesize. 

\begin{table}
\caption{Ablation results of our method on the \emph{Leaves} and \emph{Spaces$\_$073} scenes. 
We report the PSNR, SSIM, and LPIPS results between the synthesized views and the ground truth views. Our method NeRF-MIR is actually NeRF+PERE+$\mathcal{L}_{DW}$+PIRE.}
\label{tab:AblationT}
\centering
\footnotesize
\setlength{\tabcolsep}{1pt}
\newcolumntype{Y}{>{\centering\arraybackslash}X}
\begin{tabularx}{0.999\linewidth}{l||YYY|YYY}
\toprule
& \multicolumn{3}{c}{\scshape Spaces\_073 }
 & \multicolumn{3}{c}{\scshape Leaves  }

\\
  & \multicolumn{1}{c}{\scriptsize PSNR$\uparrow$}    & \multicolumn{1}{c}{\scriptsize SSIM$\uparrow$} & \multicolumn{1}{c}{\scriptsize LPIPS$\downarrow$} & \multicolumn{1}{c}{\scriptsize PSNR$\uparrow$} & \multicolumn{1}{c}{\scriptsize SSIM$\uparrow$} & \multicolumn{1}{c}{\scriptsize LPIPS$\downarrow$}  \\
  \midrule
 \scriptsize \textit{NeRF}               & 20.65 & 0.77 & 0.43    & 18.82    &  0.70    &  0.34         \\
\scriptsize \textit{NeRF+PERE} & 21.03 & 0.79 & 0.41    & 21.06    &  0.80    &  0.25
          \\
\scriptsize \textit{NeRF+$\mathcal{L}_{DW}$+PERE}              & 23.91 & 0.86 & 0.33     &  21.66    &0.81     & 0.24           \\
\textcolor{blue}{\scriptsize \textit{NeRF+$\mathcal{L}_{DW}$+PIRE}}             & 24.22 & 0.87 & 0.33     &  21.95    & 0.81     & 0.23          \\
\scriptsize \textit{NeRF-MIR(ours)}         & \textbf{26.55} & \textbf{0.90} & \textbf{0.26}   &  \textbf{23.27}    & \textbf{0.85}    & \textbf{0.22}   \\

\midrule
\bottomrule
\end{tabularx}
\vspace{-0.4cm}
\end{table}

\subsection{Ablation Study}

\noindent \textbf{Effectiveness of different components.}
We perform an ablation study on the \emph{Leaves} and \emph{Spaces\_073} scenes to evaluate the effectiveness of the key components in our approach. 

The quantitative ablation results are presented in Tab.~\ref{tab:AblationT}, as discussed in Sec.~\ref{sec:entropy}, the purpose of PERE is to efficiently capture intricate details. We can see notable performance enhancements by comparing NeRF (the 2nd row) with NeRF+PERE (the 3rd rwo), and NeRF+$\mathcal{L}_{DW}$+PIRE (the 5th row) with NeRF-MIR (the 6th row) across all metrics. Concretely, 
NeRF-MIR (the 6th row) achieves an averaged 7.8\% improvement over NeRF+$\mathcal{L}_{DW}$+PIRE (the 5th row) in PSNR, showing PERE's role in enhancing ray distribution for texture learning. 
Our proposed loss $\mathcal{L}_{DW}$ is able to exploit the masked regions at different iterations, 
by dynamically adjusting relative weights between masked regions and non-masked regions. This is validated by comparing the results of NeRF+$\mathcal{L}_{DW}$+PERE (the 4th row) and that of NeRF+PERE with the original loss in Eq.~(\ref{eq:nerf-loss})  (the 3rd row).

\begin{figure*}[t]
    \centering
    \includegraphics[width=\linewidth]{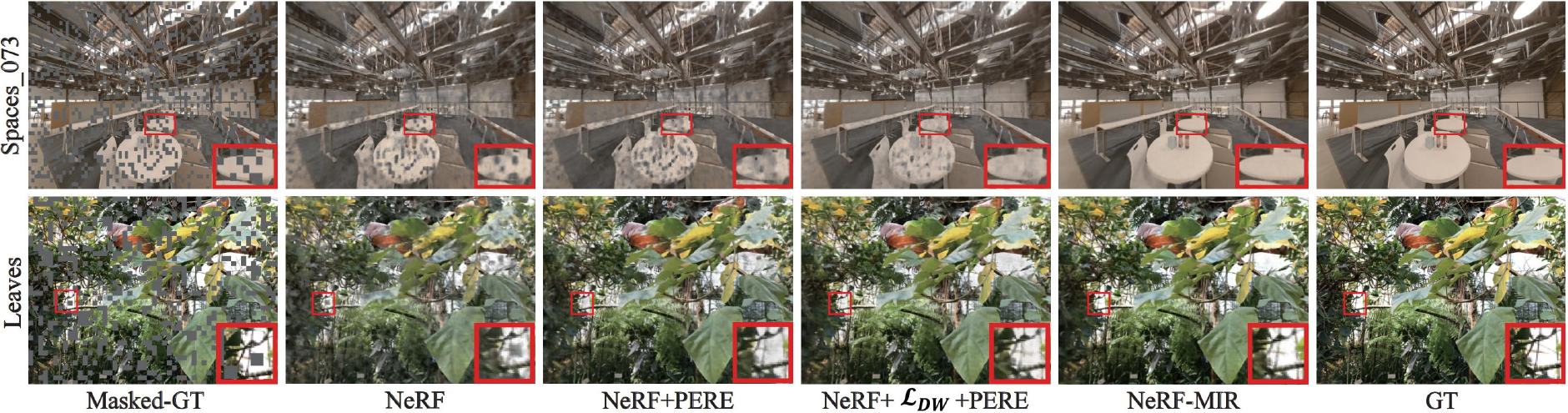}
     \caption{Ablation results of our NeRF-MIR on the \emph{Leaves} (25\%-fixed) and \emph{Space\_073} (25\%-random) scenes. From left to right, we add the PERE, $\mathcal{L}_{DW}$ and PIRE to the vanilla NeRF method one by one. Our components contribute significantly to the final restoration quality.}
   \label{fig:Ablation}
\end{figure*}

Fig.~\ref{fig:Ablation} illustrates some qualitative results, which indicate that the PERE component can significantly mitigate the impact of the masked regions, enhancing overall image quality by concentrating rays on the intricate textures.
On the other hand, the PIRE mechanism ensures the complete elimination of black dots within the highlighted red boxes, as textures in masked regions are progressively updated. 
Best performance is achieved when PERE, PIRE, and $\mathcal{L}_{DW}$ are synergistically integrated within our proposed method. 

\noindent \textbf{Robustness of mask size and masked level.}
In this part, we evaluate the robustness of mask size, examining four square mask sizes: $5 \times 5$, $10 \times 10$, $15 \times 15$ and $20 \times 20$. 
We demonstrate the effect of mask size on the \emph{Fern} scene of LLFF-M dataset in Tab.~\ref{tab:The effect of mask region size}. There are minor variations in PSNR, SSIM and LPIPS metrics as we change the mask size by $5 \times 5$, $10 \times 10$, $15 \times 15$ and $20 \times 20$, respectively.
Note that the size of masks is inversely proportional to GPU memory. Consequently, both the mask size and patch size are set at $10 \times 10$ in our experiments.

\begin{figure*}
    \centering
    \includegraphics[width=\linewidth]{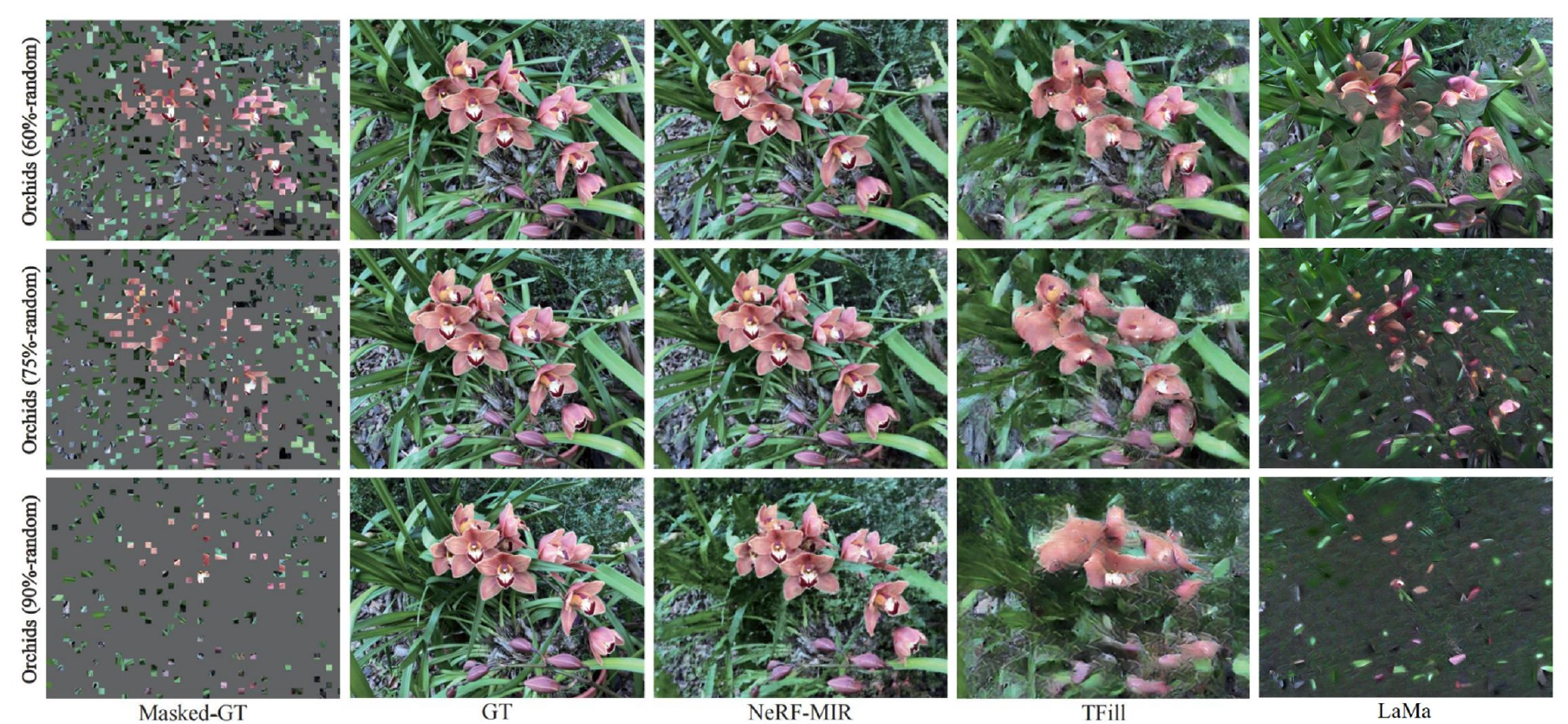}
    \caption{Restoration quality comparison at different masked levels (60\%, 75\%, 90\%) between TFill, LaMa and our method on the \emph{orchids} scene.}
    \label{fig:mask percentage}
    \vspace{-0.4cm}
\end{figure*}

We additionally investigate the impact of masked levels and compare NeRF-MIR with the learning-based approach TFill~\cite{zheng2022bridging} and LaMa~\cite{suvorov2022resolution} on different masked levels.
Fig.~\ref{fig:mask percentage} shows the qualitative results of masked image restoration. As the masked level increases,
TFill fails to reproduce the details in images, which are heavily blurred. Moreover, the appearance is inconsistent with the ground truth. Similarly, LaMa barely recovers the correct structure and texture of the target scene in high mask ratio settings.
However, our method exhibits the capability to restore the entire scene even when the masked regions encompass 90\% of the image. This shows the ability of NeRF-MIR to achieve high-quality reconstructions, even under substantial corruption. The results in Tab.~\ref{tab:Comparison of masked regions at different percentage} show that our method achieves the best results in all metrics.


\begin{table}
\caption{The effect of mask size on the \emph{Fern} scene of the LLFF-M dataset. }
\label{tab:The effect of mask region size}
\centering
\footnotesize
\setlength{\tabcolsep}{1pt}
\newcolumntype{Y}{>{\centering\arraybackslash}X}
\begin{tabularx}{0.999\linewidth}{l||YYY}
\toprule

  \textsl{Size} & \multicolumn{1}{c}{\scriptsize PSNR$\uparrow$}    & \multicolumn{1}{c}{\scriptsize SSIM$\uparrow$} & \multicolumn{1}{c}{\scriptsize LPIPS$\downarrow$}   \\
  \midrule
\textit{\scriptsize 5}           & 28.23   & 0.92    & 0.14             \\
 \scriptsize \textit{10}               & 28.21       & 0.93    & 0.13               \\
\scriptsize \textit{15}               & 28.18       & 0.93    & 0.12         \\
 \scriptsize \textit{20}               &  28.49       &  0.93    &  0.12        \\

\midrule
\bottomrule
\end{tabularx}

\end{table}


\begin{table}
\caption{Quantitative result comparison on \emph{Orchids} scene of LLFF-M between TFill and our method. The best results are in black bold. 
}

\label{tab:Comparison of masked regions at different percentage}
\centering
\footnotesize
\setlength{\tabcolsep}{1pt}
\newcolumntype{Y}{>{\centering\arraybackslash}X}
\begin{tabularx}{0.999\linewidth}{l||YYY|YYY|YYY}
\toprule
& \multicolumn{3}{c}{\scshape 60\% Mask  }
 & \multicolumn{3}{c}{\scshape 75\% Mask }
 & \multicolumn{3}{c}{\scshape 90\% Mask }
\\
  & \multicolumn{1}{c}{\scriptsize PSNR$\uparrow$}    & \multicolumn{1}{c}{\scriptsize SSIM$\uparrow$} & \multicolumn{1}{c}{\scriptsize LPIPS$\downarrow$} & \multicolumn{1}{c}{\scriptsize PSNR$\uparrow$} & \multicolumn{1}{c}{\scriptsize SSIM$\uparrow$} & \multicolumn{1}{c}{\scriptsize LPIPS$\downarrow$} & \multicolumn{1}{c}{\scriptsize PSNR$\uparrow$} & \multicolumn{1}{c}{\scriptsize SSIM$\uparrow$} & \multicolumn{1}{c}{\scriptsize LPIPS$\downarrow$}   \\
  \midrule
\scriptsize \textit{TFill}               & 16.85     & 0.62   & 0.36     &  15.10   & 0.50    & 0.45     & 12.98   & 0.34    & 0.56         \\

Ours              & \textbf{21.19}       & \textbf{0.80}    & \textbf{0.22}     & \textbf{19.57}    & \textbf{0.73}    & \textbf{0.28}     & \textbf{15.83}   & \textbf{0.51}    & \textbf{0.45}           \\
\midrule
\bottomrule
\end{tabularx}
\end{table}

\begin{figure}
   \centering
   \includegraphics[width=\linewidth]{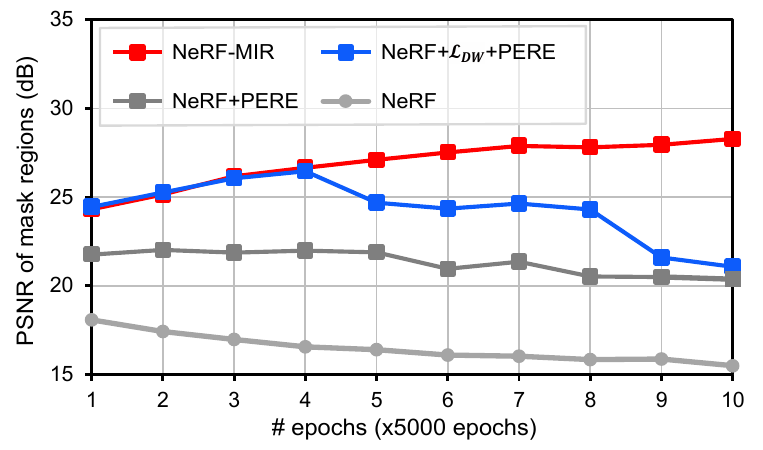}
   \caption{PSNR results for different configurations (NeRF, NeRF+PERE, NeRF+$\mathcal{L}_{DW}$+PERE and NeRF-MIR) on LLFF-M dataset. 
   }
   \label{fig:Curve}
\end{figure}

\noindent \textbf{Further analysis on the effectiveness of PERE and PIRE.} 
We further demonstrate the effects of PERE and PIRE in Fig.~\ref{fig:Curve} by comparing the variations in PSNR when training NeRF, NeRF+PERE, NeRF+$\mathcal{L}_{DW}$+PERE and NeRF-MIR. The performance of NeRF gradually degrades as the number of epochs increases, due to the fact that false information in the masked regions continuously interferes with the NeRF model. 
PSNR is significantly boosted with the inclusion of PERE, suggesting that improving ray distribution helps mitigate the influence of corrupted pixels and enhances texture recovery.
The performance is further improved with $\mathcal{L}_{DW}$+PERE. 
It is noteworthy that PSNR initially rises because the correct information from the unmasked regions dominates over the false information from masked regions. However, as iteration goes, the uncertain RGB values in masked regions gain dominance, resulting in performance degradation. 
The dynamically weighted loss function inherently acts as a regularization constraint, effectively preventing error accumulation in the PIRE process. Overall, 
our NeRF-MIR exhibits a performance-rising trend, showcasing the collaborative effectiveness of PERE, dynamically weighted loss, and the PIRE component.

\begin{figure}
   \centering
   \includegraphics[width=\linewidth]{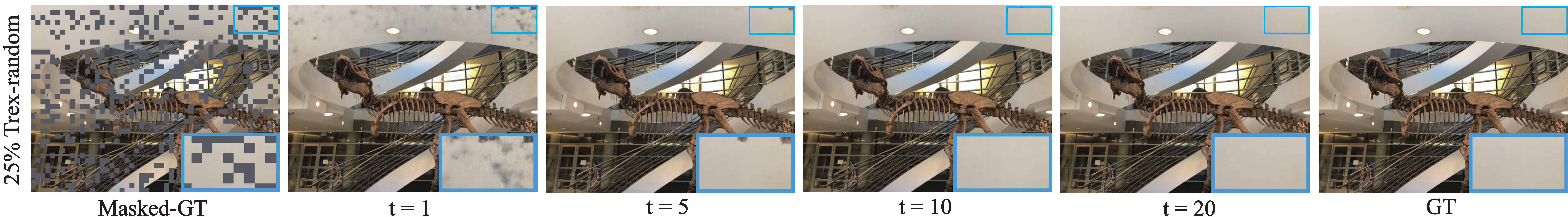}
   \caption{Qualitative results of different progressive iterations. The ground truth was masked by 25\%-random squares. We visualize the output results of $t$ with 1, 5, 10 and 20 in a consistent view. Masked areas are enlarged by blue boxes.}
   \label{fig:different number of t}
\end{figure}

\noindent \textbf{Effect of the number of iterations.}
In this part, we assess the impact of varying the value of $t$ in PIRE, comparing the changes in metrics over 100,000 training epochs by setting $t$ to 1, 5, 10 and 20, respectively. 
We present the qualitative and quantitative results in Fig.~\ref{fig:different number of t} and Tab.~\ref{tab:change of t}. 
Notably, with only one iteration, there are still visible black dots. When $t$ is set to 5, the number of black dots within the blue boxes is significantly decreased. Complete elimination of the black dots is achieved when $t$ is set to 10.
Quantitative results demonstrate that metrics of PSNR, SSIM and LPIPS are nearly invariant when $t$ are set to 5, 10 and 20, respectively.

\begin{table}
\caption{The effect of $t_i$ on the \emph{trex} scene of the LLFF-M dataset.}
\label{tab:change of t}
\centering
\footnotesize
\setlength{\tabcolsep}{1pt}
\newcolumntype{Y}{>{\centering\arraybackslash}X}
\begin{tabularx}{0.999\linewidth}{l||YYY}
\toprule

  \textsl{Iteration} & \multicolumn{1}{c}{\scriptsize PSNR$\uparrow$}    & \multicolumn{1}{c}{\scriptsize SSIM$\uparrow$} & \multicolumn{1}{c}{\scriptsize LPIPS$\downarrow$}   \\
  \midrule
\textit{\scriptsize 1}           & 28.99   & 0.95    & 0.11             \\
 \scriptsize \textit{5}               & 30.03       & 0.96    & 0.08               \\
\scriptsize \textit{10}               & 30.08      & 0.96    & 0.08         \\
 \scriptsize \textit{20}               &  30.05       &  0.96    &  0.09        \\

\midrule
\bottomrule
\end{tabularx}
\end{table}

\begin{table}[tb!]

\begin{center}
\small
\begin{tabular}{lccc}
\toprule
Component (image size)     &  GPU (GB) &
 \multicolumn{1}{c}{Runtime (s)} \\
\midrule

Masking ($378 \times 504$)   & - & 1.008  \\ 
Masking ($756 \times 1008$)   & - & 3.856  \\ 
PERE ($378 \times 504$)   & - & 0.309    \\ 
PERE ($756 \times 1008$)  & - & 1.107    \\ 
\midrule
 PIRE ($378 \times 504$) & 8 & 9.306 \\
 PIRE ($756 \times 1008$) & 16  & 37.283 \\
\bottomrule
\end{tabular}
\end{center}

\caption{The average running time of mask process, PERE and PIRE for rendering a view with the LLFF dataset. 
}
\label{table:runtime}
\end{table}

\subsection{Discussions}
\label{sec:Discussion}
\noindent \textbf{Computational cost.}
{Here, we evaluate the efficiency of the major components of our method on a personal computer equipped with a 12th Gen Intel(R) Core(TM) i5-12490F processor (3000 MHz, 6 cores, 12 logical processors). We report GPU memory consumption, the additional time incurred by PERE and PIRE when rendering a view. The results are in Tab.~\ref{table:runtime}. We can see that 
masking and calculating the entropy of a $378 \times 504$ image require an average of 1.008 and 0.309 seconds, respectively. And almost a triple time is required to process a $756 \times 1008$ image. For PIRE, each $378 \times 504$ image takes an average of 9.306 seconds to render and update, while a four-fold time is required to process a $756 \times 1008$ image.}

The overhead introduced by PIRE is primarily due to the iterative refinement process, while the cost of the offline rendering process is within an acceptable scope. Moreover, the training cost introduced by PIRE is also acceptable compared to the performance gain (11\%-15\%). The dynamically-weighted loss incurs slight overhead as it is computed along with the standard reconstruction loss.

\begin{figure*}[t]
    \centering
    \includegraphics[width=1\linewidth]{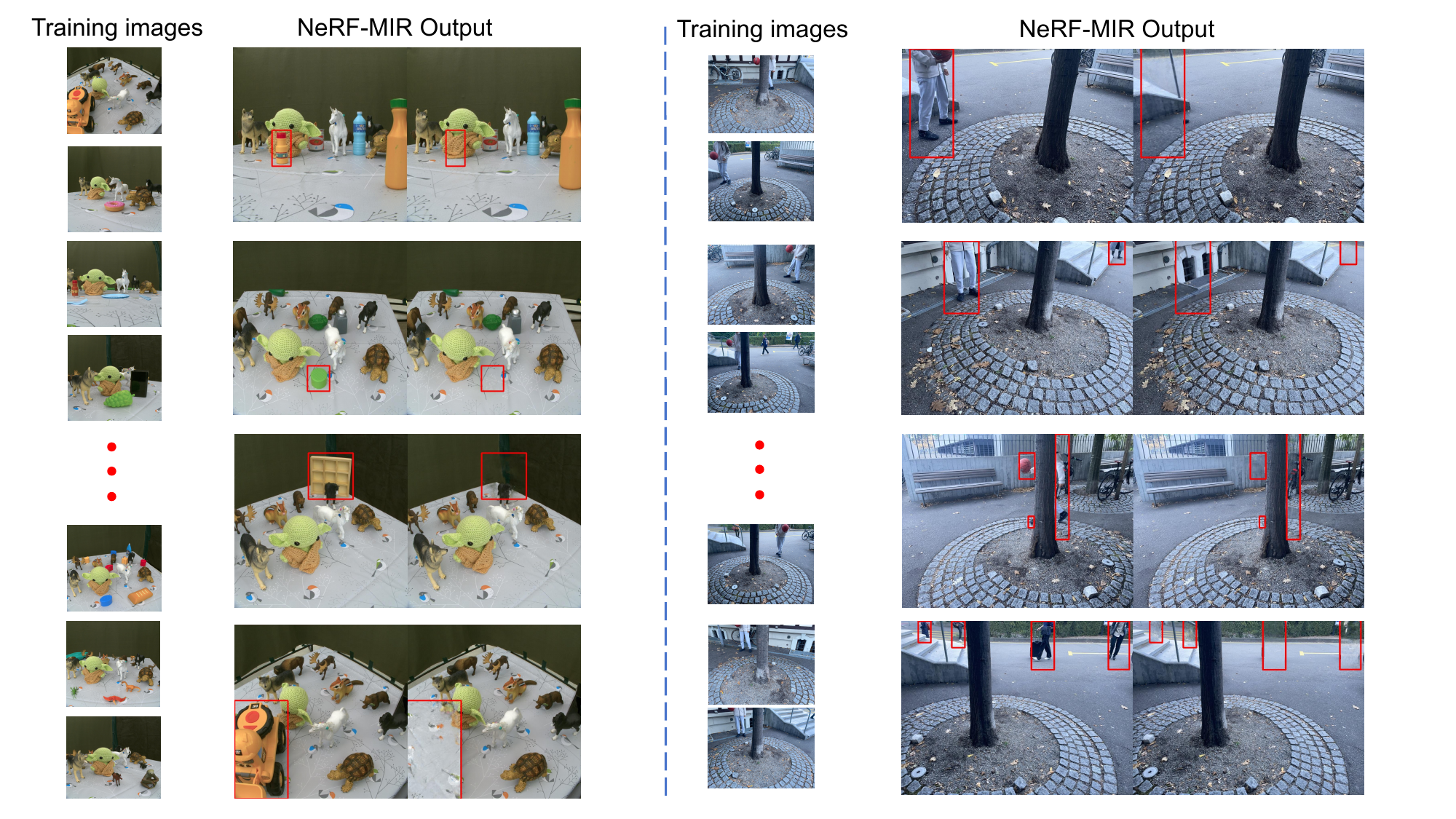}
    \caption{{The results of applying our method to the cluttered \emph{yoda} scene and on-the-go \emph{tree} scene. Here, we visualize the results of NeRF-MIR in four different views.}}
    \label{fig:application_figure}
\end{figure*}

\noindent \textbf{Applications.}
Our NeRF-MIR demonstrates its outstanding ability to be leveraged in image inpainting and disturbed scene restoration tasks. NeRF-MIR includes but is not limited to removing any type of interferer, which has a huge impact on subsequent research in this field. We summarise four potential applications for NeRF-MIR as follows:
1) \textbf{Filtering rain and snow on autopilot.} It is difficult to recovery high-quality scene from terrible circumstances~\cite{halimeh2009raindrop}(e.g. Blizzard and rainstorm). Existing target detection algorithms~\cite{bossu2011rain,wang2017hierarchical} are sophisticated enough to detect and mask these scattered occlusions. Our method is capable of synthesizing realistic results from a masked perspective captured in autonomous driving scenario~\cite{zhang2023gobigger,jia2024bench2drive,yang2025rawdrive}.
2) \textbf{Inpainting defective images or videos.} There are numerous works~\cite{bhavsar2012towards,zarif2013static} on inpainting due to sensor or lens damage. The scenarios contain unwanted artifacts, which can be attributed to sensor/lens damage or occlusions. In such a case, all the captured images contain consistent missing regions that are stationary with respect to the image coordinate system. Our method demonstrates outstanding capabilities in utilizing geometric structures
3) \textbf{Improving the integrity of training data.} NeRF-MIR can also be utilized to improve data integrity for downstream classification and detection tasks~\cite{ma2025fine,ma2025ms,zhong2025slerpface}.
4) \textbf{Enhance the existing NeRFs.} Our method seamlessly integrates with the majority of existing NeRF approaches, enhancing both their performance and convergence speed.
5) {\textbf{Clutter and pedestrian removal.} Our method can successfully remove clutter and unintentionally photographed pedestrians from on-the-go capture images. Some results are illustrated in Fig.~\ref{fig:application_figure}.}

\begin{figure}[t]
    \centering
    \includegraphics[width=1\linewidth]{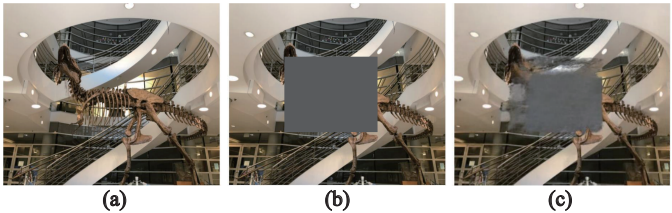}
    \caption{Failure cases on \emph{trex} with 25\%-fixed large square. NeRF-MIR can not restore the large fixed regions. (a), (b) and (c) are the GT, Masked GT and the corresponding generated results, respectively.}
    \label{fig:fail_case}
\end{figure}

\noindent \textbf{Limitations and failure cases.}
In realistic scenarios, our approach is limited by the performance of target detection methods. We will explore more advanced detection algorithms to enable NeRF-MIR to perform superior results in real-world scenarios. On the other hand, Exploring ways to detach from reliance on detection algorithms is a fascinating approach to automatic interference removal.
On the other hand, NeRF-MIR encounters challenges when a substantial region in each view is masked extensively. We demonstrate the failure cases in Fig.~\ref{fig:fail_case}. NeRF-MIR struggles to restore the large fixed regions as there is insufficient information within the masked areas. 


\noindent \textbf{Future works.}
To address the limitations of NeRF-MIR, we have developed the following future works: we intend to incorporate pose estimation into NeRF or utilize non-posed camera methods, and we believe that this will enable NeRF-MIR to recover disturbed images or videos from synthetic scenarios. 
On the other hand, 
The main drawback of NeRF-MIR is its runtime, which is a common issue with NeRFs. Nevertheless, recent advances in speeding up and parallelizing radiance field calculations(e.g., Instant-NGP, 3D Gaussian splatting) have yielded impressive results. Therefore, we will explore effective methods to reduce computational expenses, which we believe is a promising direction.

\section{Conclusion}
\label{sec:conclusion}
This paper presents the first method, named NeRF-MIR, for reconstructing 3D scenes from masked images. To enhance the sampling efficiency of rays during training, we develop a ray-emitting strategy for NeRF based on patch-based image entropy. This strategy, which determines the sampling frequency, enables NeRF-MIR to meticulously recover details in masked scenes.
A self-training approach is designed to restore the masked images progressively. In addition, we propose a dynamically weighted loss to further enhance the restoration capacity of NeRF-MIR. 
Experiments on LLFF-M, Spaces-M, Blender-M and realistic datasets demonstrate that NeRF-MIR is superior to the existing NeRF-based methods in restoring masked images. 
Additionally, we successfully remove snowflakes and defoliation in realistic scenes, which demonstrates the effectiveness of NeRF-MIR in real-world distractor removal.




\bibliographystyle{IEEEtran}
\bibliography{bibliography}

\end{document}